\pdfoutput=1
\documentclass[10pt,twocolumn,letterpaper]{article}

\usepackage{btas}
\usepackage{times}
\usepackage{epsfig}
\usepackage{graphicx}
\usepackage{epsfig}
\usepackage{graphicx}
\usepackage{amsmath}
\usepackage{amssymb}
\usepackage{algorithm,algpseudocode}
\usepackage{amsmath}
\usepackage{amsfonts}
\usepackage{amssymb}
\usepackage{enumitem}
\usepackage{subcaption}
\usepackage{dblfloatfix}
\usepackage{balance}

\floatname{algorithm}{Procedure}

\btasfinalcopy 

\ifbtasfinal\fi

\begin{document}
%
\title{Detecting Finger-Vein Presentation Attacks Using 3D Shape \& Diffuse Reflectance Decomposition}
%
%

\author{Jag Mohan Singh ~ Sushma Venkatesh ~ Kiran B. Raja \\
Raghavendra Ramachandra ~ Christoph Busch \\
{Norwegian Biometrics Laboratory, NTNU, Norway} \\
{\texttt{\{jag.m.singh; sushma.venkatesh;kiran.raja\}@ntnu.no}}\\
{\texttt{\{raghavendra.ramachandra;  christoph.busch\}@ntnu.no}}\\
}
\maketitle
\begin{abstract}
Despite the high biometric performance, finger-vein recognition systems are vulnerable to presentation attacks (aka., spoofing attacks). In this paper, we present a new and robust approach for detecting presentation attacks on finger-vein biometric systems exploiting the 3D Shape (normal-map) and material properties (diffuse-map) of the finger. Observing the normal-map and diffuse-map exhibiting enhanced textural differences in comparison with the original finger-vein image, especially in the presence of varying illumination intensity, we propose to employ textural feature-descriptors on both of them independently. The features are subsequently used to compute a separating hyper-plane using Support Vector Machine (SVM) classifiers for the features computed from normal-maps and diffuse-maps independently. Given the scores from each classifier for normal-map and diffuse-map, we propose sum-rule based score level fusion to make detection of such presentation attack more robust. To this end, we construct a new database of finger-vein images acquired using a custom capture device with three inbuilt illuminations and validate the applicability of the proposed approach. 
The newly collected database consists of 936 images, which corresponds to 468 bona fide images and 468 artefact images. We establish the superiority of the proposed approach by benchmarking it 
with classical textural feature-descriptor applied directly on finger-vein images. The proposed approach outperforms the classical approaches by providing the Attack Presentation Classification Error Rate (APCER) \& Bona fide Presentation Classification Error Rate (BPCER)  of 0\% compared to comparable traditional methods. 
\end{abstract}

\section{Introduction}
The use of finger-vein as a type of a biometric characteristic for verification is growing in recent years, especially for banking transaction applications, where there is a need for highly secure access control \cite{Hitachi2004}. The practical example of such systems can be found in commercial applications that were developed by Hitachi in 2004 with finger-vein verification mainly for use in unsupervised scenarios such as Bank ATMs \cite{Hitachi2004}. 
Despite the high recognition accuracy of such finger-vein systems, they are prone to attacks at the capture level where one can use an artefact (e.g., printed finger-vein image) to gain illegitimate access, especially in unsupervised access control settings. Such attacks at the capture level are popularly termed as presentation attacks (aka., spoofing attacks), and attacks can be carried out using different kind of artefacts, such as using a printed photo of a finger-vein representation (print-attack) or alternatively using an electronic display to show the vein representation (display-attack)\cite{Tome14}.
\begin{figure}[h!]
\centering
\includegraphics[width=1.0\linewidth]{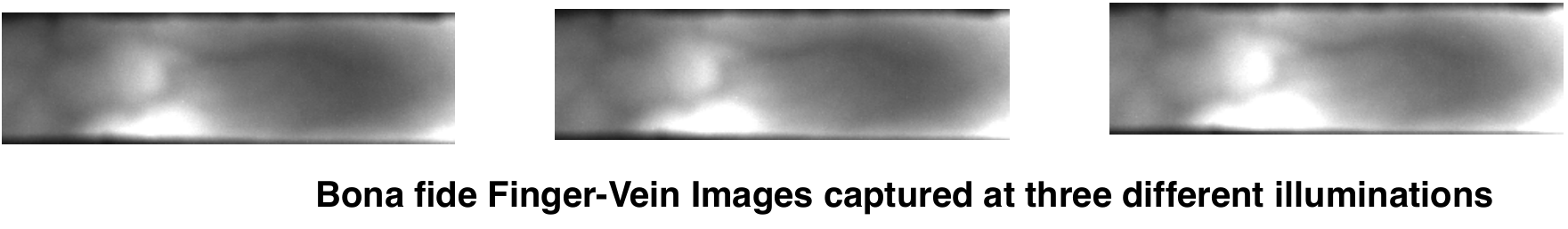} 
 \caption{Bona fide finger-vein image captured in three different illumination intensities.}\label{fig:SingleFingerVeinImage}
\end{figure}

\begin{figure*}[b]
\centering
\includegraphics[width=0.95\linewidth]{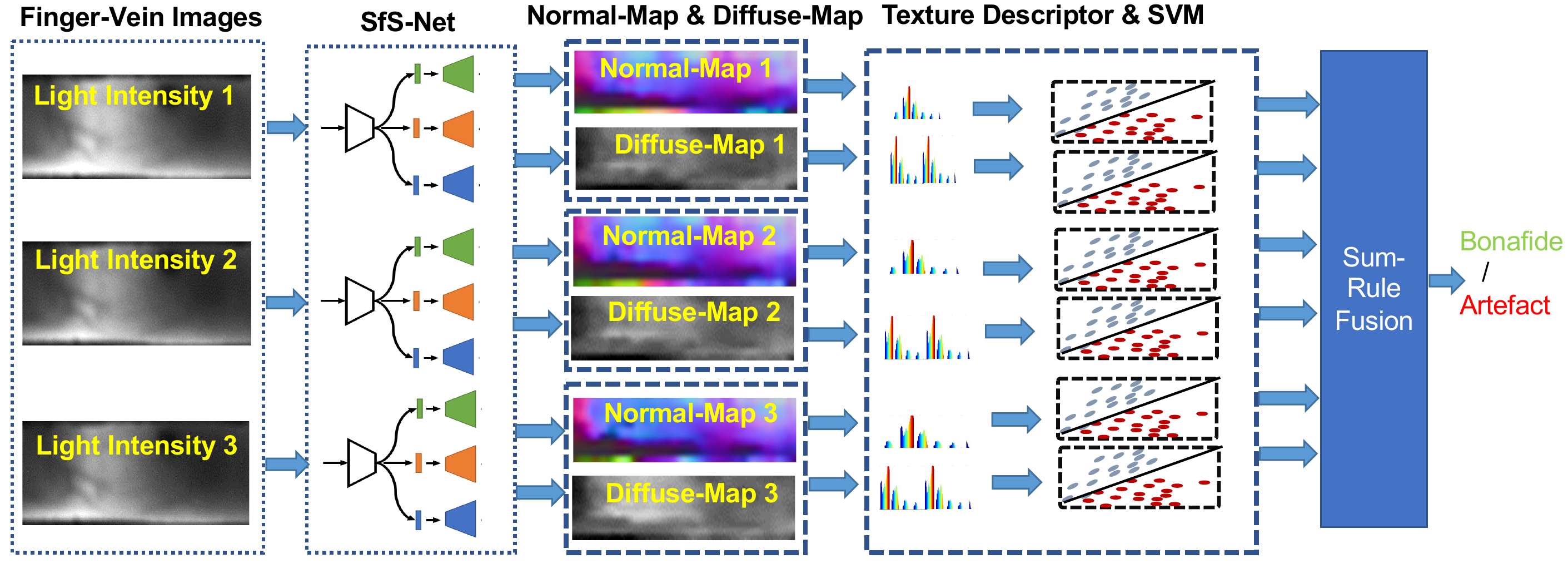}
 \caption{Block Diagram of proposed approach for finger-vein presentation attack detection.}\label{fig:ProposedAlgorithmPAD}
\end{figure*}

The vulnerability of finger-vein verification systems to presentation attacks was first shown by Nguyen et al. in 2013~\cite{Nguyen13}. Given the vulnerability finger-vein verification systems to presentation attacks, it is essential to increase their security by supplementing with robust attack detection mechanisms~\cite{China18}. 

In the rest of the paper, we present the related work, and our contributions in Section~\ref{related}, followed by the proposed approach in Section~\ref{proposedalgorithm}, followed by Finger-Vein capture device, and database in Section~\ref{device}, and experimental protocol \& results in Section~\ref{results}. In the end, conclusions \& future-work are presented in Section~\ref{conclusion}.

\section{Related Work \& Our Contributions}\label{related}
Nguyen et al.~\cite{Nguyen13} proposed the first PAD system for finger-vein based on Fourier and Wavelet transforms. 
In 2015 the first finger-vein PAD competition was organized~\cite{Tome15}, the participating teams proposed the following main feature extraction methods: Binarised Statistical Image Features (BSIF~\cite{BSIFPaper}), and a combination of a set of local texture descriptors including Local Binary Patterns (LBP~\cite{LBPPaper}), Local Phase Quantization (LPQ~\cite{LPQPaper}), a patch-wise Short Time Fourier Transform (STFT), and a Weber Local Descriptor(WLD) each of them were followed by SVM classifier which resulted in highly accurate performance. Further, in 2015, Raghavendra et al.~\cite{Raghavendra15BTAS} used a Eulerian Video Magnification method to quantify the blood-flow for finger-vein PAD against print-attacks. They benchmarked their approach with state-of-the-art (SOTA) presented in~\cite{Tome15}, achieving significant performance improvement over them. A new Presentation Attack Instrument (PAI) based on electronic display was later proposed by Raghavendra et al.~\cite{Raghavendra15SITIS}, where steerable pyramids were used for finger-vein PAD. Tirunagari et al.~\cite{Tirunagari15} presented a novel approach where they used Windowed Dynamic mode decomposition (W-DMD) on static images of finger-vein instead of image sequences as done previously in Dynamic mode decomposition. Kocher et al.~\cite{Kocher16} introduced LBP extensions for finger-vein PAD and concluded that they performed the same as the "LBP" baseline. In deep-learning-based approaches, Qiu et al.~\cite{Qiu17}  designed a new convolutional neural network (CNN) called FPNet for finger-vein PAD, which achieved perfect accuracy over VERA DB~\cite{VERADB}, which is a public database. Nguyen et al.~\cite{Nguyen17} developed a PAD method for finger-vein recognition systems based on capture from a near-infrared (NIR) camera. They extracted image features using a CNN followed by principal component analysis (PCA) for dimensionality reduction and a Support Vector Machine (SVM) as a classifier. Raghavendra et al.~\cite{Raghavendra18Database} presented the augmented pre-trained CNN models that have demonstrated outstanding performance on both print and electronic display attack. Qiu et al.~\cite{Qiu18} used total variation decomposition to divide the finger-vein sample into the structural and noise components to achieved a D-EER of 0\% and used an image decomposition approach. Maser et al.~\cite{Maser19} used Photo Response Non-Uniformity (PRNU) to detect PAD for finger-vein images and demonstrated their results on two publicly available datasets IDIAP~\cite{Tome14} and SCUT-FVD~\cite{Qiu18}. Anjos et al.~\cite{Anjos19} have summarized the efforts in finger-vein presentation attacks, and their detection.
\subsection{\bf{Our Contributions}}
In summary, we note that the use of textural information in a classifier can be well used for PAD in finger-vein systems. We employ the custom-built finger-vein sensor that can capture the finger-vein characteristics under three different illumination intensities.
On the decomposed normal-map and diffuse-map of the finger-vein image, we extract the textural-descriptors in line with the previously published works. Thus, with the given approach of decomposed normal-map and diffuse-map, it is possible to employ any of the existing finger-vein PAD approaches. The proposed approach is presented further in detail in Section~\ref{proposedalgorithm}. In our proposed approach, we also work on minimizing the pre-processing and manual effort required in finger-vein extraction algorithms. We mainly don't have the finger-vein extraction step in our proposed approach, which involves a lot of manual effort as the finger-vein extraction process requires the tuning of parameters.
The key contributions of this work, therefore, can be summarized as:
\begin{itemize}
     \item Presents a novel approach of using material properties (diffuse-map) and 3D Shape (normal-map) directly on the captured finger-vein image and is fully backward compatible with previous feature descriptor based approaches.
    \item Our approach does not require extensive manual effort, and pre-processing, which is one of the main drawbacks of existing methods. The manual effort is needed in two stages for many existing approaches, wherein the first stage is the finger-vein to be visible in the captured image. In the second stage, the finger-vein has to be extracted from the captured image. Our approach only requires the finger-vein to be partially visible in the captured image, and the second stage of finger-vein extraction is not needed in our proposed approach.
    \item Presents a baseline evaluation with feature descriptors applied directly on the finger-vein images, and the same with feature descriptors post decomposition into diffuse-map and normal-map.
    \item Presents an extensive evaluation of the proposed approach on a newly collected database of $936$ images from $78$ unique subjects.
\end{itemize}

\section{Proposed approach}\label{proposedalgorithm}
In this section, we describe the proposed approach for robust finger-vein PAD. The proposed approach is based on computing 3D shape (normal-map) and material properties (diffuse-map) from a single finger-vein image. We assert that the 3D characteristics differ for bona fide and artefact finger-vein image samples. The key reason for this difference is attributed to the 3D shape (normal-map), and material properties (diffuse-map) being different for natural (bona fide) finger-vein, and an artefact of the same finger-vein. This difference is further highlighted by varying the illumination intensity, which makes it suitable for PAD. The idea of using illumination intensities is relatively new and was pointed by authors in~\cite{Tai18}, where it was pointed out that the usually ignored factors of illumination intensity and exposure would play a role in capturing the 3D Shape from a semi-calibrated photometric stereo. Thus, we use varying illumination intensities to capture the 3D Shape in multiple illumination intensities, as it is much easier to vary the illumination intensity of the compared to changing its direction. Figure~\ref{fig:ProposedAlgorithmPAD} shows the block diagram of the proposed approach that can be structured in four functional units that are explained as below: 
\subsection{{\bf{Finger-Vein Image Capture}}} As illustrated in
Figure~\ref{fig:SingleFingerVeinImage}, we capture the finger-vein from the sensor using three illuminations with varying illumination intensities. The three illumination intensities are chosen for each finger such that the best quality finger-vein image-based is obtained by the human operator for \textit{intensity-1} and subsequently, the images are obtained for illumination \textit{lower} than \textit{intensity-1} and illumination \textit{higher} than \textit{intensity-1}. We indicate these three images from three intensities as $I_{1}$, $I_{2}$, and $I_{3}$.
\begin{figure}[h!]
\centering
\includegraphics[width=1.0\linewidth]{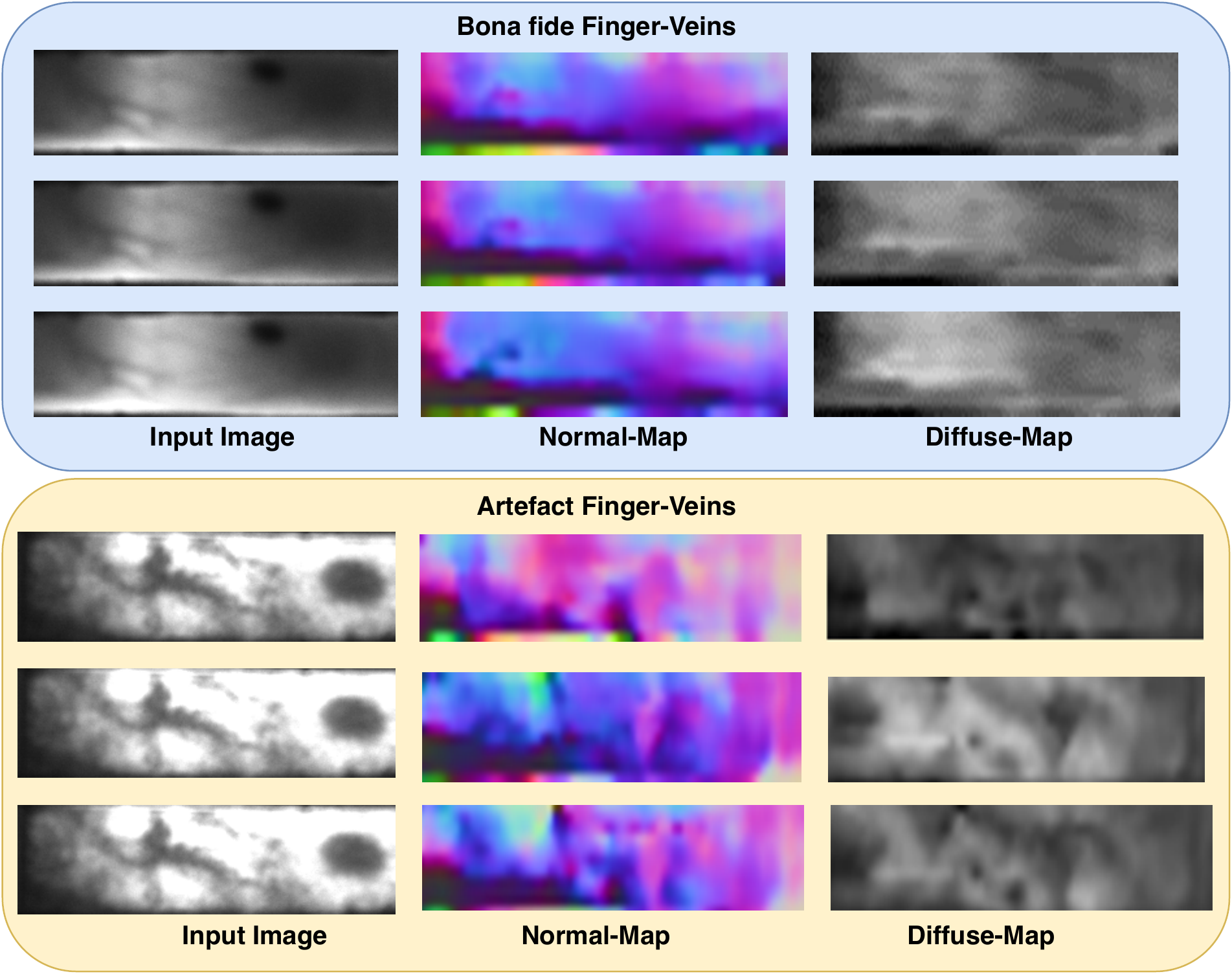}
\caption{Decomposition of bona fide, and artefact images into normal-map, and diffuse-map.}\label{fig:Decomposition}
\end{figure}

\subsection{{\bf{Normal-Map \& Diffuse-Map Decomposition}}} Given the three finger-vein images captured ($I_{1}$, $I_{2}$, and $I_{3}$), we obtain normal-map (indicated by $I_{1}^{N}$, $I_{2}^{N}$, and $I_{3}^{N}$) and diffuse-map (indicated by $I_{1}^{D}$, $I_{2}^{D}$, and $I_{3}^{D}$) for each of three finger-vein images for a subject.  Assuming a Lambertian reflectance model of the underlying surface (either bona fide or artefact), we extract normal-map and diffuse-map using SfS-Net~\cite{Soumyadip18}. It can be further noted that normal-map and diffuse-map provide textural features that can be employed for PAD\footnote{\scriptsize{Although SfS-Net~\cite{Soumyadip18} provides diffuse-map, normal-map, albedo-map, and shading-map, we notice usable features for PAD in diffuse-map and normal-map alone.}}. 
\begin{algorithm}[h!] 
\caption{Normal-Map \& Diffuse-Map Decomposition}
\label{alg:sfsnetdecomp}
\begin{algorithmic}[1]
\Require{$I_{1}, I_{2}, I_{3}$} 
\Ensure{$I_{1}^{D}$, $I_{2}^{D}$, $I_{3}^{D}$, $I_{1}^{N}$, $I_{2}^{N}$, $I_{3}^{N}$}
\Procedure{\footnotesize{Compute Normal-Map and Diffuse-Map}}{$I_{1}, I_{2}, I_{3}$}
   \For{$k\gets 1,2,3$}
   \State Compute Normal-Map ($I_{k}^{N}$) directly from SfS-Net.
    \State Compute Diffuse-map ($I_{k}^{D}$) for $I_{k}$ by Equation~\ref{eq1}, and Equation~\ref{eq2}.
    \EndFor
    \State \Return {$I_{1}^{D}$, $I_{2}^{D}$, $I_{3}^{D}$, $I_{1}^{N}$, $I_{2}^{N}$, and $I_{3}^{N}$}
    \EndProcedure
\end{algorithmic}
\end{algorithm}
The diffuse-map ($I_{k}^{D}$) and normal-map for $I_{k}^N$ can easily be obtained using second-order spherical harmonic lighting coefficients under the assumption of Lambertian reflectance model for $k = {1, 2, 3}$ corresponding to three images of finger-vein images. We obtain shading-map ($r(I_{k}^{N}(p))$) of the material for $k^{th}$ input finger-vein image (using notation from~\cite{Basri03}) using Equation~\ref{eq1} where $l_{nm}$ are spherical harmonic lighting coefficients, and $r_{nm}$ are the harmonic images.
We then obtain the diffuse-map ($I_{k}^{D}$) using albedo-map ($\rho_{k}$), and shading-map obtained in Equation~\ref{eq1} using Equation~\ref{eq2}.
The decomposition of the input finger-vein images into diffuse-maps, and normal-maps is presented as an Algorithm~\ref{alg:sfsnetdecomp}.
\begin{align}
  r(I_{k}^{N}(p)) = \sum_{n=0}^{n=2} \sum_{m=-n}^n l_{nm} r_{nm}((I_{k}^{N}(p)) \label{eq1}\\
  I_{k}^{D} = \rho_{k} r(I_{k}^{N}(p))\label{eq2}
 \end{align}

One can observe from  Figure~\ref{fig:Decomposition}, the textural difference in bona fide and artefact finger-vein presentations are emphasized post decomposition in both normal-map and diffuse-map. Further, with the lighting estimation using spherical harmonics, we also note that artefacts result in enhanced noise for extracted normal-map and diffuse-map~\cite{SHNoise}. The enhanced noise would result in better detection of presentation artefacts.
\begin{figure}
    \includegraphics[width=0.5\textwidth]{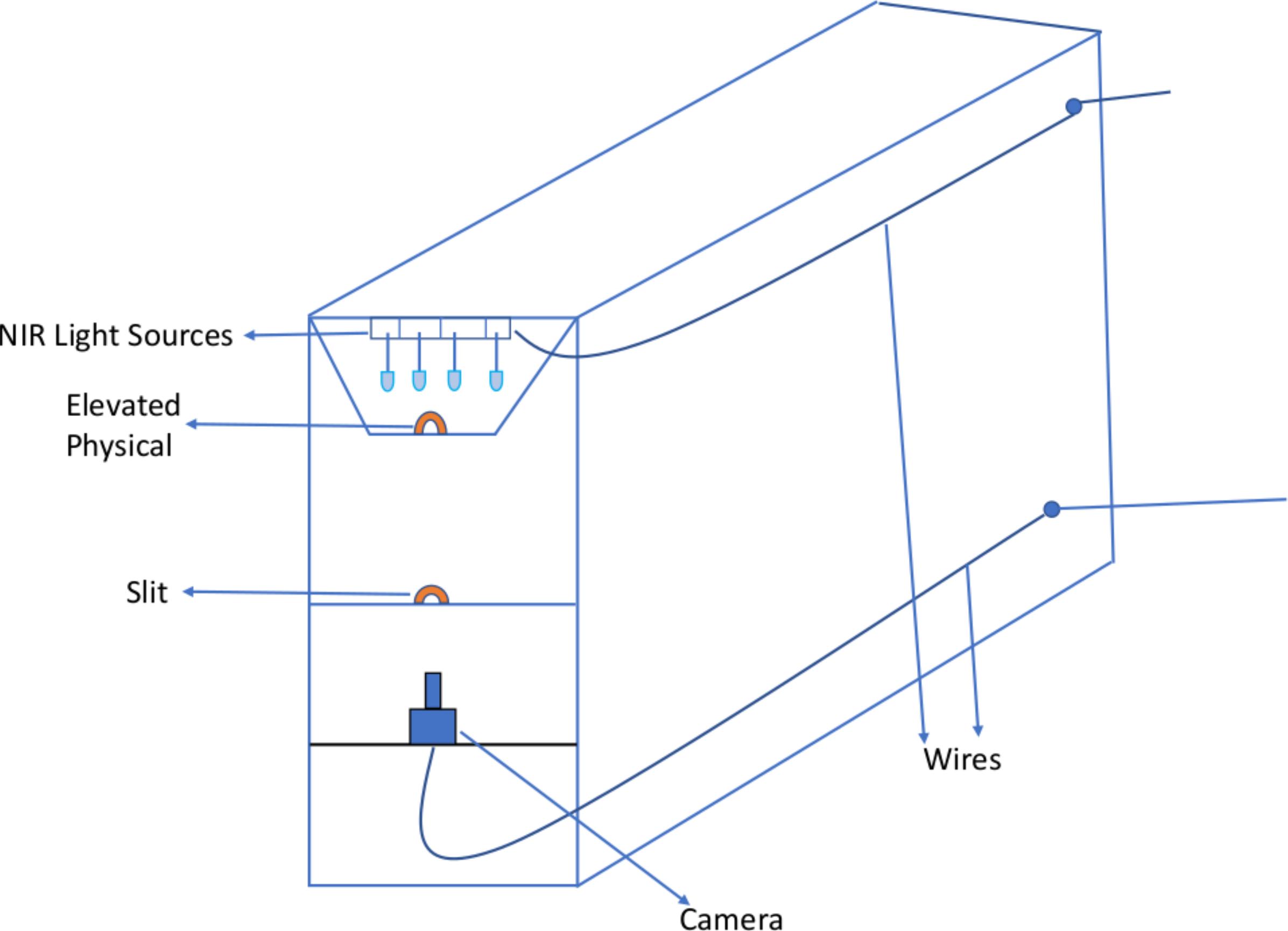}
\caption{Schematic description of our finger-vein capture device. It is used capture finger-vein images in 3 different illumination intensities.}\label{fig:SchematicSensor}
\end{figure}

\begin{figure*}
    \centering
    \includegraphics[width=0.75\textwidth]{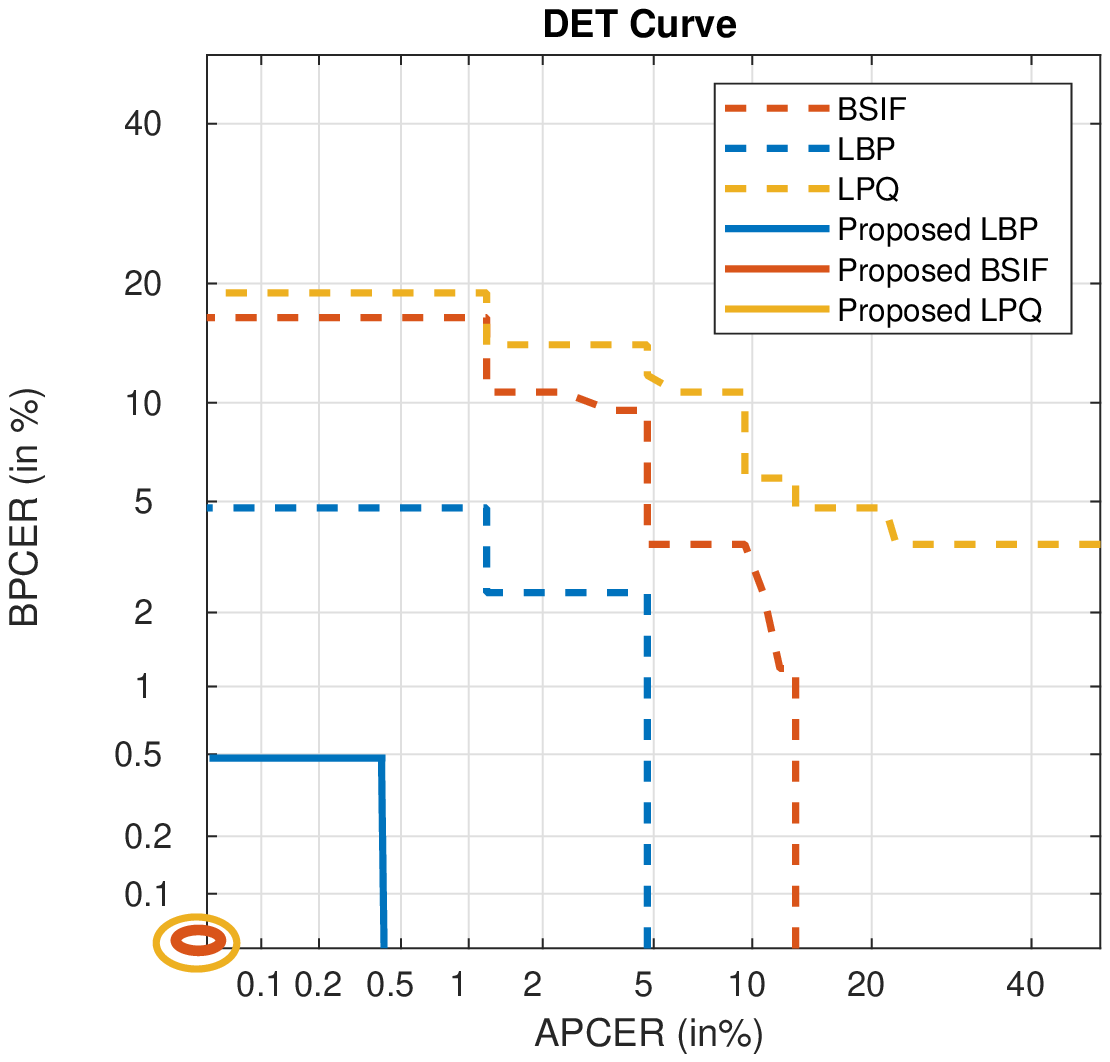}
\caption{DET Curves for LBP, BSIF, and LPQ (w and w/o proposed approach) where initial arcs show $0$ EER for (proposed approach+BSIF, and proposed approach+LPQ) and 0.5 EER for proposed approach+LBP}\label{fig:DETCurves}
\end{figure*}

\begin{table*}[htb]
\centering
\resizebox{1\linewidth}{!}
{\begin{tabular}{|c|c|c|c|c|c|c|c|c|c|}
  \hline
    {\bf{Method}} & \multicolumn{3}{|c|}{\bf{D-EER}} & \multicolumn{3}{|c|}{\bf{BPCER@}} & \multicolumn{3}{|c|}{\bf{BPCER@}} \\ 
    
          & \multicolumn{3}{|c|}{}    & \multicolumn{3}{|c|}{\bf{APCER=5\%}} & \multicolumn{3}{|c|}{\bf{APCER=10\%}} \\ 
          \cline{2-10}
         & $I_{1}$ & $I_{2}$ & $I_{3}$  & $I_{1}$ & $I_{2}$ & $I_{3}$ & $I_{1}$ & $I_{2}$ & $I_{3}$ \\ \hline 
{LBP-SVM} & \bf{16.6} &   \bf{3.5} & \bf{1.19} & \bf{51.1} & \bf{2.3} & \bf{0} &  \bf{34.5} & \bf{0} & \bf{0} \\ \hline
{LPQ-SVM} & \bf{5.9} & \bf{8.3} & \bf{16.6} & \bf{5.9} & \bf{11.9} & \bf{20.2} & \bf{3.5} & \bf{7.1} & \bf{19.0} \\ \hline
{BSIF-SVM} & \bf{11.9} & \bf{4.7} & \bf{9.5} & \bf{26.1} & \bf{4.7} & \bf{9.5} & \bf{16.6} & \bf{4.7} & \bf{9.5} \\ \hline
{Proposed + LBP (Normal-Map)} & \bf{7.1} & \bf{8.3} & \bf{11.9} & \bf{8.3} & \bf{17.8} & \bf{15.4}  & \bf{3.5} & \bf{3.5} & \bf{11.9} \\ \hline
{Proposed + LBP (Diffuse-Map)} & \bf{0} & \bf{1.1} & {1.1} & \bf{0} & \bf{0} &  \bf{0} &\bf{0} & \bf{0} & \bf{0}  \\ \hline
{Proposed + LPQ (Normal-Map)} & \bf{1.1} & \bf{1.1}  & \bf{1.1} & \bf{0} & \bf{0} & \bf{0} & \bf{0} & \bf{0} & \bf{0} \\ \hline
{Proposed + LPQ (Diffuse-Map)} & \bf{0} &  \bf{0} & \bf{0} & \bf{0} & \bf{0} & \bf{0} & \bf{0} & \bf{0} & \bf{0} \\ \hline
{Proposed + BSIF (Normal-Map)} & \bf{2.3} & \bf{2.3}  & \bf{2.3} & \bf{0} & \bf{0} & \bf{0} & \bf{0} & \bf{0} & \bf{0} \\ \hline
{Proposed + BSIF (Diffuse-Map)} & \bf{0} &  \bf{0} & \bf{0} & \bf{0} & \bf{0} & \bf{0} & \bf{0} & \bf{0} & \bf{0} \\ \hline
\end{tabular}}
\caption{Proposed approach with Illumination Intensity 1 ($I_{1}$), Illumination Intensity 2 ($I_{2}$), and Illumination Intensity 3 ($I_{3}$) results for diffuse-map and normal-map.}\label{Results1}
\end{table*}

\subsection{\bf{Feature Extraction \& Classification}} Given the set of normal-maps and diffuse-maps of the finger-vein images, we observe the differences in textural properties corresponding to bona fide finger-vein and artefact finger-vein images as illustrated in Figure~\ref{fig:Decomposition}. To this extent, we employ LBP, BSIF, and LPQ to extract the features of normal-map and diffuse-map. We then train a linear SVM classifier for both the normal-maps and diffuse-maps separately based on lighting. Further, with the availability of three independent scores for normal-map and three independent scores for diffuse-map, we employ score level \textit{SUM-Rule} fusion to obtain a final PAD score, which is used for classification. 

\section{Finger-Vein Capture Device \& Database}\label{device}
\subsection{Finger-Vein Capture Device}
Given the unavailability of databases for the validation of our proposed approach, we construct a custom finger-vein database using a low-cost custom-built finger-vein capture device, whose schematic is shown in Figure~\ref{fig:SchematicSensor}.
Our finger-vein capture device has four important parts, as follows:
\begin{itemize}
    \item NIR Light Sources.
    \item Camera and Lens. 
    \item Elevated Physical Structure.
    \item Slit.
\end{itemize}
We use 20 LEDs of 980 nm for the near-infrared (NIR) light source, and industrial camera DMK 22BUCO3 with a resolution of $744\times480$ pixels, and T3Z0312CS lens with focal length 8mm is used. Elevated Physical Structure is used to obtain a sufficient amount of light for the image, and slit is for the placement of the finger. To obtain the finger-vein images, the correct amount of light must penetrate through the finger. The correct amount of light is achieved by placing a physical structure with elevation, which helps in the concentration of light. The user places the finger in the slit, which is a small gap between NIR light sources and the camera. The placement of NIR Light Source is that finger-veins are blocked while imaging from the camera in which that finger-vein patterns appear as dark patterns in the finger-vein image obtained.

\begin{table*}[thp]
\centering
\resizebox{1\linewidth}{!}
{\begin{tabular}{|c|c|c|c|}
  \hline
    {\bf{Method}} & \bf{D-EER} & \bf{BPCER@} & \bf{BPCER@} \\ 
          &     & \bf{APCER=5\%} & \bf{APCER=10\%} \\ \hline
{LBP~\cite{LBPPaper} \& SVM (Sum-Rule Fusion)} & 2.3 & 0 & 0  \\ \hline
{LPQ~\cite{LPQPaper} \& SVM (Sum-Rule Fusion)} & 9.5 & 10.7 & 5.9  \\ \hline
{BSIF~\cite{BSIFPaper} \& SVM (Sum-Rule Fusion)} & 4.7 & 3.5 & 3.5  \\ \hline
{Proposed approach + LBP \& SVM (Sum-Rule Fusion)} & {\bf{0.5}} & {\bf{0}} & {\bf{0}} \\ \hline
{Proposed approach + LPQ \& SVM (Sum-Rule Fusion)} & {\bf{0}} & {\bf{0}} & {\bf{0}} \\ \hline
{Proposed approach + BSIF \& SVM (Sum-Rule Fusion)} & {\bf{0}} & {\bf{0}} & {\bf{0}} \\ \hline
\end{tabular}
}
\caption{LBP-SVM, BSIF-SVM, and LPQ-SVM with and without proposed approach.}\label{Results2}
\end{table*}

\subsection{Finger-Vein Database}
We first capture the bona fide finger-vein images for  $78$ unique subjects using $3$ illuminations with a resolution of $744\times480$ for each image. For each subject, the only right index finger is captured in two different sessions over a gap of 3-4 weeks. Thus, the whole dataset is comprised of $78$ Subjects with $78$ finger-vein samples captured in $3$ illumination intensity, and $2$ sessions = $468=78\times3\times2$ bona fide finger-vein samples, and same number of artefacts $468=78\times3\times2$. 
We generate the Presentation Attack Instrument (PAI) (i.e., artefacts) by printing the finger-vein images using a RICOH MP 8002 printer using a glossy printing paper. To create high-quality artefacts that can challenge the finger-vein recognition system, we first enhance the finger-vein pattern by employing the STRESS algorithm following the earlier works~\cite{Raghavendra19} before printing. We then capture the artefacts using the same capture device with three different illuminations. We divide the finger-vein dataset into training and testing partition by dividing them into non-overlapping sets of 36 subjects in training with two sessions, and 42 subjects in testing with two sessions. Thus, in training set corresponding to first illumination intensity, we generate $72=36\times2$ bona fide features, and $72=36\times2$ attack features for diffuse-maps, and normal-maps. Similarly, for illumination intensities two, and three, we generate $72=36\times2$ bona fide features, and $72=36\times2$ attack features for diffuse-maps, and normal-maps. Now, for the testing set for illumination intensity one, we generate $84=42\times2$ bona fide scores, and $84=42\times2$ attack scores for both diffuse-maps, and normal-maps. The same number of testing scores are generated for illumination intensities two, and three, which are $84=42\times2$ bona-fide scores, and $84=42\times2$ attack scores for both diffuse-maps, and normal-maps.

\section{Experimental Results}\label{results}
For the set of experiments conducted in this work, we report the PAD performance using the metrics defined in the International Standard ISO/IEC 30107-3~\cite{ISO2017}. We also report Detection Equal Error Rate (D-EER $\%$) and detection error trade-off curves.

Table~\ref{Results1} shows results in detail for each light, and normal-map and diffuse-map separately where diffuse-map is performing better. Diffuse-Map has more textural differences than normal-map. Additionally, it captures the difference in material properties of bona fide finger-vein (skin) v/s artefact (printed paper in our case). It can be noted from Table~\ref{Results1} that with LPQ, $0$ D-EER is achieved for diffuse-map, and normal-map. However, LBP on individual normal-maps gives results close to the LBP-SVM applied directly on the finger-vein image. Table~\ref{Results2} presents the results of the proposed approach obtained by sum-rule fusion and compares it with classical texture-descriptor based state-of-art approaches.   As it can be noted from the Table~\ref{Results2}, the proposed approach outperforms existing SOTA using texture-descriptors, we achieve best D-EER of $0.0\%$ compared to best D-EER of SOTA of $2.3\%$. The results can also be seen in Figure~\ref{fig:DETCurves}, which presents the Detection Error Trade-off Curves illustrates that the sum-rule fusion of scores results in a higher level of performance.

\section{Conclusions \& Future-Work}\label{conclusion}
In this paper, we presented a approach using single image decomposition into normal-map, and diffuse-map of a finger-vein image results in higher accuracy as compared to directly using texture-descriptors on it. Our technique does not require elaborated pre-processing (image enhancement), and manual methods (finger-vein extraction). We have further validated our proposed approach on a newly collected finger vein-image database. Future work will include the evaluation of the proposed approach on a large scale database.
\section*{Acknowledgement}This work was carried out under the funding of the Research Council of Norway under Grant No. IKTPLUSS 248030/O70. 
\balance
{
\bibliographystyle{ieee}
\bibliography{submission_example}
}
\end{document}